\documentclass[10pt,twocolumn,letterpaper]{article}
\usepackage[bookmarks=false]{hyperref}
\usepackage{wacv}
\usepackage{times}
\usepackage{epsfig}
\usepackage{graphicx}
\usepackage{amsmath}
\usepackage{amssymb}
\usepackage{booktabs}
\usepackage{caption}

\wacvfinalcopy 

\def\wacvPaperID{1141} 

\begin{document}

\title{End to End Lip Synchronization with a Temporal AutoEncoder}

\author{Yoav Shalev \hspace{2cm} Lior Wolf\\
School of Computer Science\\Tel Aviv University\\
}

\maketitle

\begin{abstract}
   We study the problem of syncing the lip movement in a video with the audio stream. Our solution finds an optimal alignment using a dual-domain recurrent neural network that is trained on synthetic data we generate by dropping and duplicating video frames. Once the alignment is found, we modify the video in order to sync the two sources. Our method is shown to greatly outperform the literature methods on a variety of existing and new benchmarks. As an application, we demonstrate our ability to robustly align text-to-speech generated audio with an existing video stream. Our code and samples are available at https://github.com/itsyoavshalev/End-to-End-Lip-Synchronization-with-a-Temporal-AutoEncoder.
\end{abstract}

\section{Introduction}

With the advancement of video technology, one may assume that lip syncing would become a thing of the past. However, with the ongoing shift form captured to generated media, coupled with a surge in user generated content apps, the need for effective methods is rapidly growing.

Consider, for example, the scenario of a generated video. It could involve the head of one actor, extracted from an old footage, the lips of another actor, added to match the script, and the voice of a Text to Speech (TTS) robot. Syncing the different sources, and especially the lip motion to the audio, to which viewers are very sensitive, poses a challenge.

As another example, consider the trending lip syncing apps. Users try their best to align their lips with a song they choose. In many cases, this alignment is only partly successful and the acquisition process needs to repeat itself.

Despite the well-defined task and the clear application need, the literature is relatively limited. Many of the existing solutions perform a global shift, which cannot address many of the modern use cases. Other methods rely on unrealistic assumptions and are only partly trained. In addition, while modifying the video without causing artifacts is much easier than modifying the audio, existing methods choose the latter, creating metallic voices and other artifacts.

In this work, we employ an end to end temporal autoencoder to perform the alignment. The method projects simple video and audio features to a joint space, where their mutual distances are considered. An LSTM autoencoder, with an added attention layer is then used to predict the correct alignment for temporal windows of up to one second. A straightforward matching method is then used to align sequences of unlimited length.

In an extensive series of experiments, we compare our method to the literature algorithms and show a sizable improvement in accuracy. Our method is robust enough to work in real-world scenarios and the vast majority of users cannot tell the difference between a video that was modified and an untouched video, even when matching to the audio track of a different speaker.

\section{Related Work}
Dynamic time warping (DTW) \cite{MULER}, \cite{10.2307/2030214} uses a dynamic programming to align multiple time-series. DTW measures the similarity and finds an optimal match by inserting frames. Anguera et al. \cite{anguera2014audio} use a predefined similarity between graphemes and phonemes. Tapaswi et al. \cite{2014arXiv1411.5726V} introduce a similarity between visual scenes and sentences based on appearance of same characters to align TV shows and plot synopses. Other dynamic programming approaches have also been used for multi modal alignment of text to speech \cite{haubold2007alignment} and video \cite{zhu2015aligning}. DTW was extended using canonical correlation analysis (CCA) for learning a mapping \cite{6126545}, \cite{zhou2012generalized}, \cite{zhou2009canonical}. CCA based DTW models aren't able to model nonlinear relationships. This has been addressed by Deep Canonical Time Warping (DCTW) ~\cite{trigeorgis2016deep}, which learns non-linear, correlated and temporally aligned representations of multiple time-series.

The earliest methods in the field of lip motion synchronization, employ phoneme recognition for the audio stream~\cite{doi:10.1002/vis.4340020404, 5745053} and align them with the mouth positions~\cite{doi:10.1002/vis.4340020404} or with visemes~\cite{5745053}, which are face templates that correspond to a specific configuration of the face during speech production. Zoric and Igor classified the MFFC representation of the audio signal into visemes and then aligned them with a parametric face model~\cite{Zoric}. 

Other methods perform the alignment without employing visemes or phonemes.  Argones et al. used co-inertia analysis and a coupled hidden Markov model~\cite{Argones}. Sargin et al. used canonical correlation analysis for speech and lip texture features~\cite{4351913}. 

Within the domain of deep learning methods, \cite{Marcheret2015DetectingAS} used a network that operates on predefined audio and visual features, in contrast to our work where we predict a shift for each frame and learn the visual features directly. Chung et al. used a CNN model called SyncNet to learn visual features directly from the video stream, and audio features from the MFCC representation of the audio~\cite{Chung2016OutOT}. They then find the optimal global shift by computing the contrastive loss over all frames. 
In contrast, we employ an alignment network and are not limited to a global shift.

In contrast to the metric learning approach of SyncNet, Chung et al. learned audio-visual embeddings as a multiclass classification task~\cite{2018arXiv180908001C}. Their method classifies one visual frame into one multiple possibly matching audio frames obtained at $N$ different locations. The decisions are integrated in order to obtain a global shift.


Realizing that in many cases a single global shift is suboptimal, Halperin et al. have shown that a network that is similar in structure to SyncNet can be leveraged to achieve a dynamic temporal alignment~\cite{2018arXiv180806250H}. This is done by using the Dijkstra  algorithm as a post processing step, in order to find the optimal path in the graph of pairwise distances.

A related task is that of reconstructing a 3D mesh animation, or generating a video of a talking face, which is aligned to an input audio signal. Karras et al. predict 3D vertex coordinates of a target face model, given a waveform of any source speaker, and aim to control the emotional states of the generated face~\cite{NVIDIA}.  ObamaNet~\cite{2018arXiv180101442K} uses a recurrent neural network (RNN) in order to predict a PCA representation of normalized mouth key-points, given a raw waveform created by a text to speech module. The key-points are then used to condition a generator that synthesizes a video of a target identity. Suwajanakorn et al. trained an RNN on many hours of Obama's weekly address footage to map from the MFCC representation of the input audio stream into mouth shapes~\cite{Suwajanakorn:2017:SOL:3072959.3073640}. Then, they synthesized a talking face video of Obama. Son Chung et al. proposed an encoder-decoder model that learns the joint embeddings of a given audio and a target face images, and then generates a lip-synced talking face video~\cite{2017arXiv170502966S}.

\section{Proposed Method}
\begin{figure*}
\includegraphics[width=\linewidth]{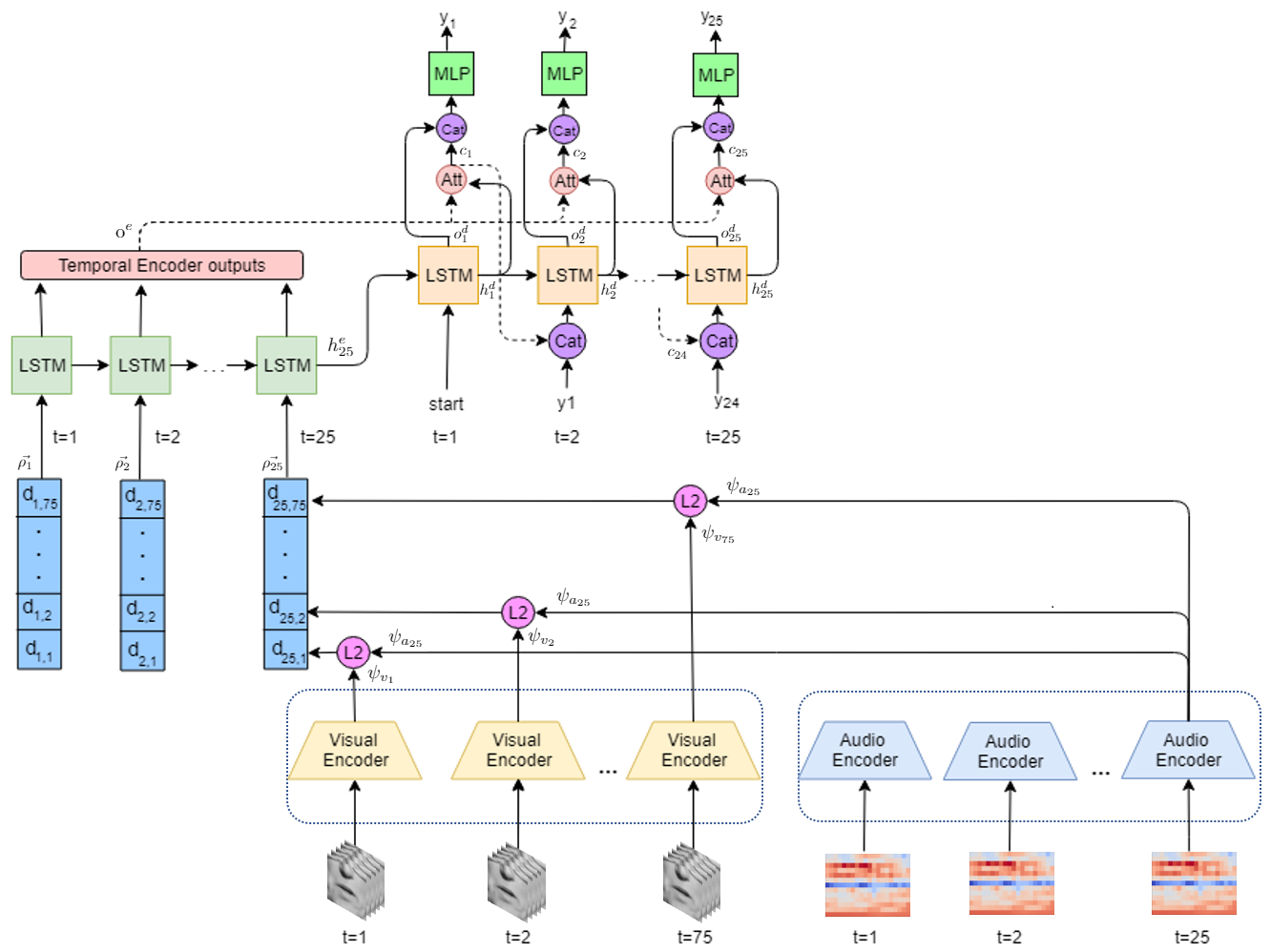}
\begin{center}
\end{center}
   \caption{Overview of our method. For every audio feature a vector of Euclidean distances is computed with all video features. The vectors are fed into an LSTM encoder followed by LSTM decoder, in order to predict a sequence of indices from the visual stream.
  }
\label{fig:system_overview}
\end{figure*}

An overview of our method is shown in Fig.~\ref{fig:system_overview}. The inputs to our model are unsynchronized audio and visual streams. The visual stream and the audio stream are both encoded to represent the local neighborhoods as tensors. A temporal encoding is obtained by considering pairwise distances between the video and the audio tensors.  
The temporal encoding then goes through an attention modulated LSTM decoder. At every time step, the decoder attends the encoder's outputs and predicts an index of the input's video stream.

\subsection{Inputs}

\begin{figure}[t]
  \centering
  \begin{tabular}{cc}
  \includegraphics[width=.2145\linewidth]{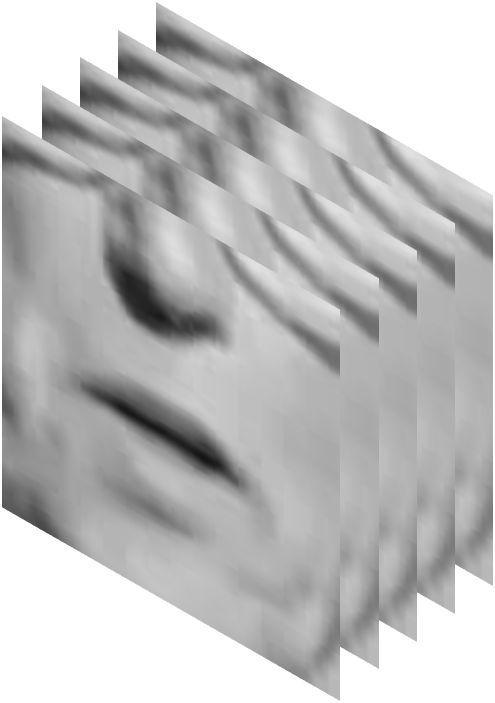}&
    \includegraphics[width=.4845\linewidth]{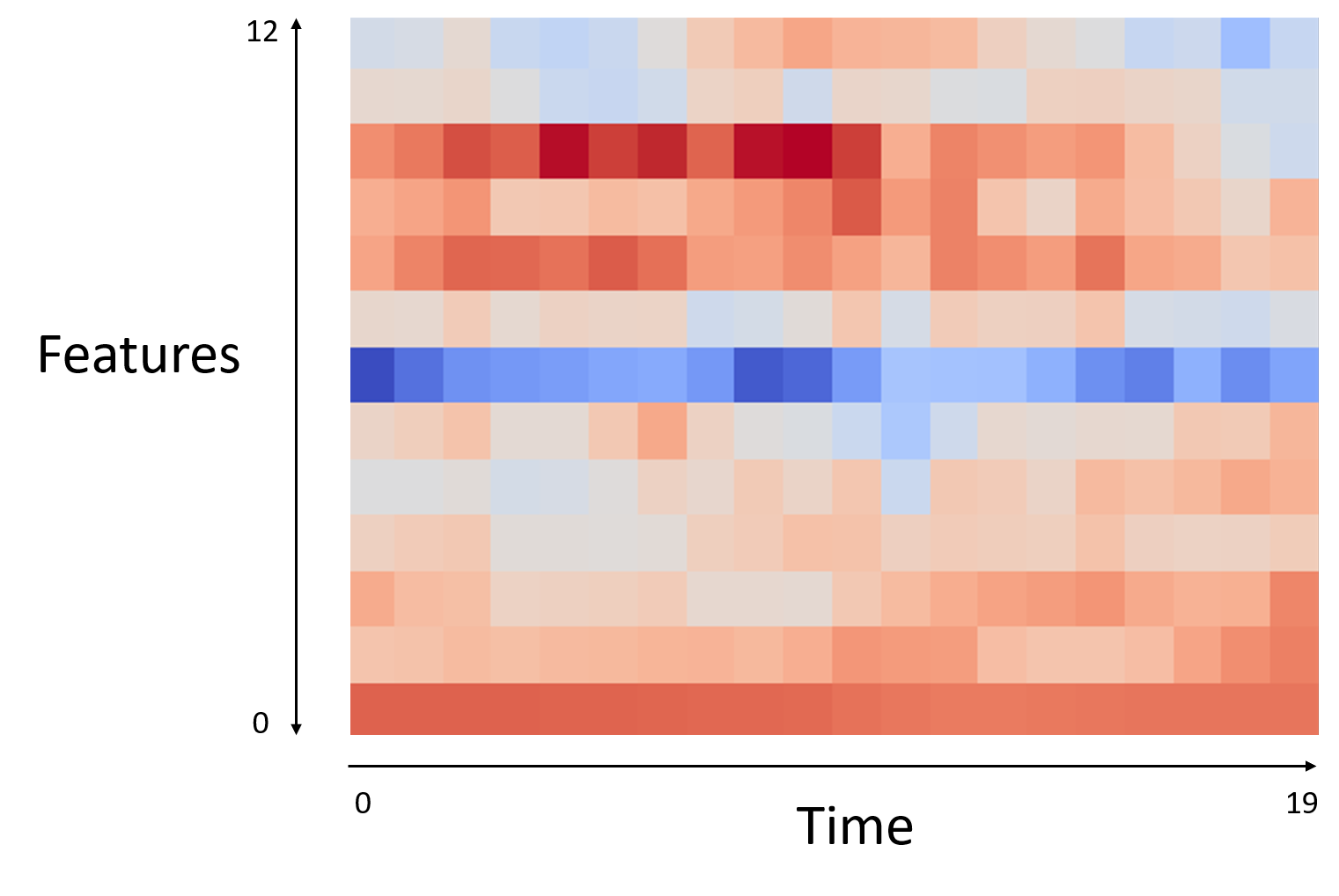}\\
    (a) & (b)\\
    \end{tabular}
  \caption{Input preprocessing. (a) Each video input is a temporal stack of five consecutive video frames of the mouth area. (b)  Each audio input consists of 20 time steps of 13 Mel-frequency bands MFCC. \label{fig:input}
  }
  
\end{figure}

As a preprocessing step for the video stream, facial key-points and rotations are extracted using Openface~\cite{Baltrusaitis2018OpenFace2F,2016arXiv161108657Z}. We then extract the mouth area, align it to the vertical axis, and normalize its size to $120\times 120$ pixels. Each video input is a temporal stack of five consecutive video frames, and the stride of consecutive video inputs is one frame, see Fig.~\ref{fig:input}(a).

The alignment network has a temporal receptive field of three seconds at 25 FPS, i.e., 75 frames. Each frame is represented by a tensor of size $5\times 120\times 120$, where the five channels correspond to the five consecutive frames above. This redundant representation, in which the same cropped mouth frames appear in multiple tensors,  follows the early fusion model suggested by Chung and Zisserman~\cite{Chung2016LipRI}.

The input audio is sampled at 16kHz. Each single audio input (``audio frame'') consists of 20 time steps of MFCC features which are corresponding to a 0.2 second input signal. Specifically, 13 Mel-frequency bands are used for the MFFC features, with a window length of 25 ms and a window step of 10ms. The stride between consecutive audio inputs is 40ms, which is the same as one video frame at 25 FPS. An example of a single audio input frame is shown in Fig.~\ref{fig:input}(b).

The network has a receptive field of one second, i.e., 25 audio frames, where each frame is represented by a matrix of size $13\times 20$, corresponding to the 20 time steps of MFCC features detailed above. 

Note that in our method, video and audio are not treated in a symmetric way, and we match a second of audio to three seconds of video. The underlying reason is that we want to allow for each audio frame a search window of 25 video frames on both sides.

\subsection{Method and Architecture}

For given audio and visual tensors of lengths $n=25$ and $m=75$, respectively, each audio and visual input is encoded using the audio or visual encoder into feature  vectors of size 1024:
\begin{equation}
\begin{split}
\psi_{a_i}=E^a(a_i) &\quad i \in \{1, ..., n\} \\
\psi_{v_j}=E^v(v_j) &\quad j \in \{1, ..., m\}
\end{split}
\end{equation}
Where $\psi_{a_i}$ ($\psi_{v_j}$) are the embeddings of audio (video) input at time step i (j). In order to emphasize our method's advantage over the literature, we employ the same architecture for the audio and video encoders ($E^a$ and $E^v$) as the one used in SyncNet~\cite{Chung2016OutOT}. This architecture consists of 3D convolutions (operating also on the temporal domain) with max-pooling layers, ending with fully connected layers. Batch normalization layers are added after each convolutional layer.

Next, for each $\psi_{a_i}$ we calculate the Euclidean distance with all $\psi_{v_j}$ and obtain the Euclidean distances features vectors of length $m$:
\begin{equation}
\begin{split}
\Vec{\rho_{i}}=[d_{i,1}, ..., d_{i,m}]\\
\end{split}
\end{equation}
where $d_{i,j}=\|\psi_{a_i}- \psi_{v_j}\|$.

The sequence of vectors $\rho = [\Vec{\rho_1}...\Vec{\rho_n}]$ is then encoded using a recurrent neural encoder that produces outputs $o^e$ and hidden states vectors $h^e$:

\begin{equation}
\begin{split}
o^e_{i}, h^e_{i} =\operatorname{RNN_e}\left(\Vec{\rho_i},h^e_{i-1}\right)\\
\end{split}
\end{equation}
and then decoded using a recurrent neural decoder followed by a multi-layer perceptron (MLP). As suggested by Bahdanau et al.~\cite{2014arXiv1409.0473B}, in order to allow the decoder to attend the entire encoded sequence at every output step, we are using an attention mechanism. At every output step $k$, the decoder produces state vectors $h^d_k$ and output vectors $o^d_k$, based on  the previous step context vector $c_{k-1}$, the decoder state vector $h^d_{k-1}$ and the predicted visual frame index $y_{k-1}$. The latter is obtained by an MLP that is applied to the output vector $o^d_{k-1}$ and the context vector $c_{k-1}$. 

Written explicitly, the RNN decoder has the following input-output structure:
\begin{equation}
\begin{split}
o_{k}^{d}, h_{k}^{d} =\operatorname{RNN_d}\left(h_{k-1}^{d}, y_{k-1}, c_{k-1}\right),
\end{split}
\end{equation}
where, the next step context vector $c_k$ is computed using the attention model:
\begin{align}
e_{k, i}=U^{T} \tanh \left(W h_{k}^{d}+V o_i^e+b\right)
\end{align}
\begin{align}
\alpha_{k, i}=\operatorname{Softmax}\left( e_{k,\cdot}\right)=\frac{\exp \left(e_{k, i}\right)}{\sum_{i'=1}^{n} \exp \left(e_{k, i'}\right)}
\end{align}
\begin{align}
c_{k}=\sum_i \alpha_{k,i}{o}^e_i 
\end{align}
where $U$, $W$, $V$ and $b$ are learned weights, and  $\alpha_{k,i}$ are the attention weights, which are obtained by a softmax operator applied to the pre-weights $e_{k,i}$.

The context vector $c_k$ is concatenated with the output vector $o^d_k$ and fed into the MLP, which outputs a vector of size $1\times m$. Then the predicted index of the visual frame that matches audio frame $k$ is given by applying a softmax layer:
\begin{align}
p = P\left(y_{k} | v, a, y_{k-1}\right)  = \operatorname{Softmax}\left( \operatorname{MLP} \left(o_{k}^{d}, c_{k} \right)\right) \label{eq:p}
\end{align}
\begin{align}
y_{k}  =  \operatorname{argmax}_i p_i
\end{align}

At $k=1$, the context vector $c_0$ is set to $0$, the input $y_0$ is set to a special symbol that indicates a start of a sequence, and the decoder state $h^d_0$ is set to the last encoder's hidden state $h^e_n$. 

The architecture of the RNN encoder and decoder consists of three LSTM layers with hidden sizes of 512. The MLP consists of two linear layers with 512 and 256 hidden units respectively, followed by a ReLU activation and another linear layer that outputs a vector of size 75.


\subsection{Loss Function}

By predicting sequences of indices of the visual stream, we are treating the problem as a multiclass classification problem. Therefore we use the cross entropy loss:
\begin{equation}
\mathcal{L}(p,t) = -\log \left(\frac{\exp (p_t)}{\sum_{j} \exp (p_j)}\right),
\end{equation}
where $p$ is the vector of assignment probabilities, given in Eq.~\ref{eq:p} and $t$ is the ground truth index of the video frame that matches the current audio frame. This loss is aggregated across all audio frames in all sequences of the training set. 

\subsection{Training Details}

In order to train the model for predicting the local frames shifts, we created a synthetic training dataset out of the LRS2-BBC dataset~\cite{2018arXiv180902108A} as described in Sec.~\ref{Datasets}. 

We employ a two phase training. In the first, we employ a version of the LRS2-BBC in which only a single shift is performed, and pretrain for 20 epochs the input encoders $E^a$ and $E^v$. The full model, including all networks, is then trained for 30 epochs, with a batch size of four, using the full-blown synthetic data set.

The LSTMs are initialized using semi-orthogonal matrices as suggested by Saxe et al.~\cite{2013arXiv1312.6120S}, the convolutional and linear layers were initialized using Xavier Normal~\cite{pmlr-v9-glorot10a} and the batch normalization layers were initialized with a normal distribution ($1 \pm 0.02$).  We use an Adam optimizer~\cite{2014arXiv1412.6980K} with a learning rate of $10^{-6}$ and beta coefficients of 0.5 and 0.999. At train time, we apply dropout with 0.1 probability on the LSTM decoder's previous-step output-embeddings.
 
In order to improve performance and reduce overfitting, we applied data augmentations for the visual input at train time. We multiplied the tensor of the visual information by a random value and randomly flipped, scaled, rotated and translated all images in the tensor.

\section{Finding Optimal Alignment at Inference}

We assume that at the starting point of the sequence, the optimal alignment distance of the video and the audio streams is up to two seconds. If the shift is larger, a global alignment step is first required. 

The learned method is applied to a second of audio and three seconds of video. At inference time we are required to align, in a coherent manner, long sequences of visual frames. 
In addition, 
a direct inference may be suboptimal, even for short sequence. Recall that at each time step, the decoder outputs a probability for each of multiple indices of the video stream. A greedy approach would predict for each audio frame the most probable visual frame index. This index, due to the decoder's structure, is conditioned on the past frames, but not on the future frames.

In order to overcome these challenges, we perform the following steps. First, we apply the method multiple times with a stride of ten frames. At each application, we employ a three second temporal window in the video domain that is centered around the one second window in the audio domain. As a result, for each audio frame we obtain up to three votes for a prediction of a video frame index. 

We treat these votes as equal and consider for each audio frame, as possible matching candidates, only the set of indices with the maximal number of votes. For example, an audio frame can be associated twice with index $i_1$ and once with index $i_2$. In this case, we only consider the index $i_1$ as a potential match. In other cases, three different indices may be potential matches.

We then perform dynamic programming in order to find the longest sequence of matches with monotonically increasing indices. In other words, the search assigns matches to one audio frame at a time, going sequentially over all audio frames. It only considers future matches that do not go back (multiple votes to the same video frame are possible), and stops the sequence, if no such further matches are possible. The length of such monotonic sequences is optimized.

At the end of the obtained sequence, there is at least one frame for which a match cannot be found without going back. We skip such frames and continue with a longest sequence search when an audio frame has a possible match that does not break monotonicity. 




In order to smooth the matching results and fill-in the missing matches, we apply an adaptive smoothing procedure as suggested by Halperin et al.~\cite{2018arXiv180806250H}. The procedure employs a Laplacian filter with a parameter $\sigma$ and a smoothness criterion. It searches for the lowest $\sigma$ such that the criterion is met. This is done with one such parameter for the entire video sequence. Predictions that are shorter than the audio stream are padded with the last visual predicted frame.

\section{Datasets} \label{Datasets}

Our network is trained on a dataset that is derived from the LRS2-BBC dataset~\cite{2018arXiv180902108A}. Another split of this dataset is used for evaluation. We also perform evaluation by switching the audio between two TCD-TIMIT~\cite{7050271} speakers uttering the same test. 

\subsection{LRS2-BBC}
LRS2-BBC consists of thousands of spoken sentences from BBC television. Videos are of frame size of $160 \times 160$, with a frame rate and audio rate of 25 FPS and 16kHz respectively. For training, we use the pre-train set, which consists of 96,318 utterances. The test set, which consists of 1,243 utterances, is used for evaluation in our experiments. 

The dataset itself is well-synchronized and we adapt it to our purposes by augmenting the data. In the video stream, frames are randomly dropped and duplicated, while maintaining a maximal distance of 25 frames from the frame's original position, which bounds the number of drops and duplications. We also restrict the augmetnation such that the maximal number of occurrences for a single frame is four. 

The target prediction is set such that if a visual frame is dropped, the corresponding audio frame is mapped to the closest remaining frame, and if a visual frame is duplicated, the corresponding audio frame is mapped to the last occurrence of that frame. The audio stream is also augmented, by adding a global random shift inside the three second visual stream window. The size of the shift is {[-25...25]} frames.

\begin{figure*}[t]
  \centering
  \includegraphics[width=.8145\linewidth]{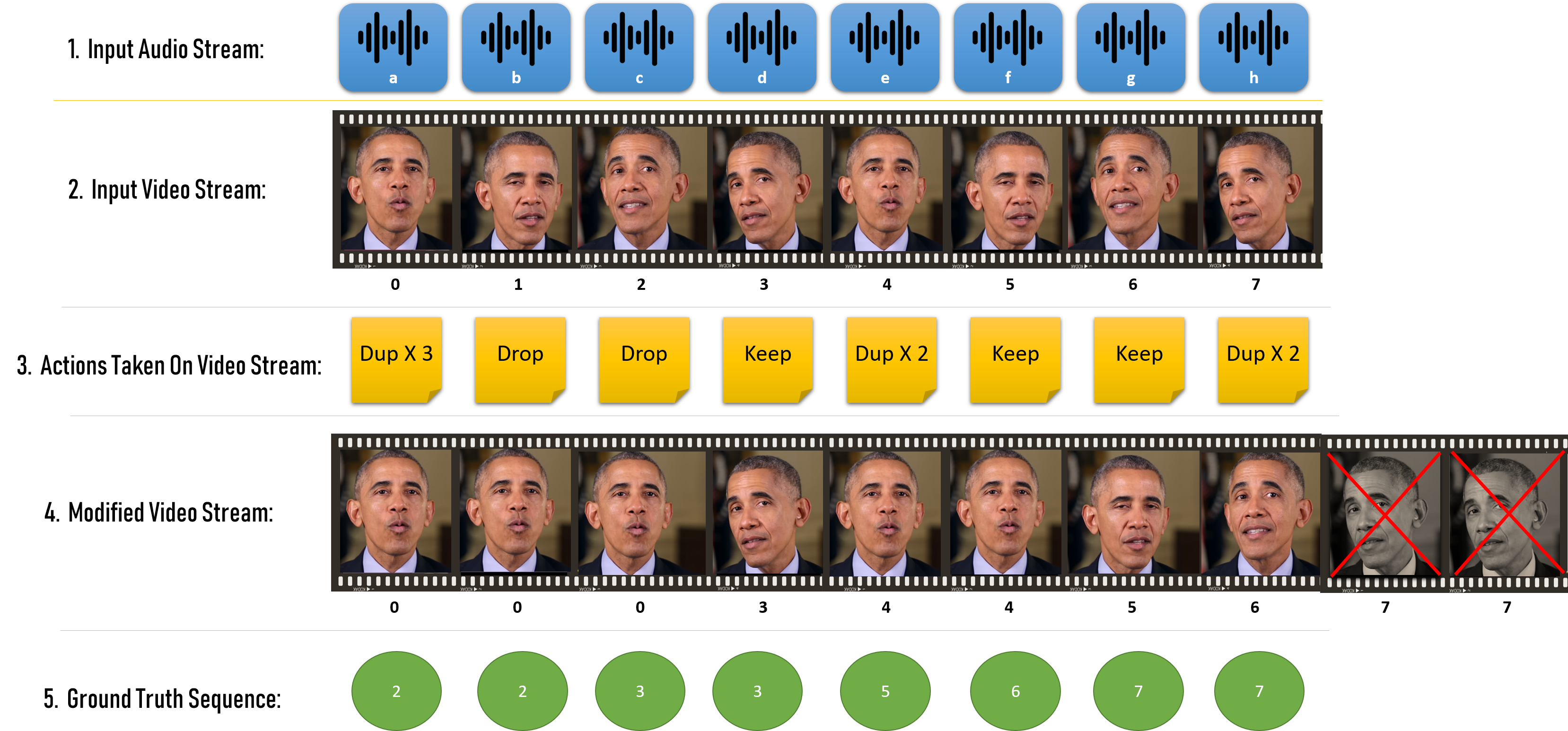}
  \caption{The creation process of a training audio-video pair, along with its ground trough labels. For a given audio-video pair (1, 2), a sequence of random actions is taken (3) on the visual stream, to create an unsynchronized video stream (4) along with the ground truth labels (5). The resulted training pair consists of the input audio (1) and the modified video stream (4) and the task is to predict the ground truth labels (5). Note that the frames at the end of the modified video stream are removed since they fall after the audio boundaries.
  }
  \label{fig:db_creation}
\end{figure*}

An illustration of the process is shown in Fig.~\ref{fig:db_creation}.

\subsection{TCD-TIMIT}
The TCD-TIMIT dataset, developed for speaker and speech recognition purposes, is used for evaluation only. The dataset consists of $1,920\times 1,080$ videos of 62 speakers, both males and females, saying 6,913 sentences. Videos are down-sampled by us into a frame rate and audio rate of 25 FPS and 16kHz respectively. 

The sentences vary between speakers. However, at the start of the session, all speakers say the same two sentences -- sa1 and sa2. For evaluation, we took sentences sa1 and sa2 of the first five speakers which are three males and two females. We then consider the various combinations of video from one speaker and speech audio from another.

\section{Experiments} 
\label{Experiments}

We compare the performance of the proposed method to existing methods for visual speech synchronization. We evaluate both in the case in which synchronization is obtained by applying a single shift for the entire video, and in the case where the shift is different for each frame. For the local shift case, we perform experiments where either the audio or the video are modified. Out of these options, we strongly advocate for a local form of synchronization and for modifying the video. Local alignment allows for a greater flexibility and generalizes the global shift. Modifying the video results in changes that are far less noticeable, while modifying the audio often results in metallic sounds and other artifacts, which are not easy to overcome. The reason that we evaluate our method with the other options is to match the literature methods, in the same domain in which these methods operate. 

A summary of the literature baselines is given in Tab.~\ref{tab:baselines}. Since no other method in the literature, as far as we know, performs local shifts and warps the video, we created a suitable baseline method by modifying the Dynamic Temporal Alignment (DTA) method of Halperin et el.~\cite{2018arXiv180806250H}. The original method is local and warps the audio. We have implemented their method for video augmentation by inverting the search of the Dijkstra matching to be from audio to video.

\begin{table}
\begin{center}
\begin{tabular}{cp{2cm}p{2cm}}
    \toprule
    Method & Global/local alignment & Augments video/audio\\
    \midrule
    Ours & both & both\\
    \midrule
    SyncNet & global & same \\
    DTA & both & audio \\
    Our modified DTA & both & video \\
    \bottomrule
    Chung et al.\cite{2018arXiv180908001C} & global & same \\
\end{tabular}
\end{center}
\caption{The literature baselines. The baseline below the ruler were not available for evaluation. For global methods, aligning the video is performed similarly to aligning the audio. \label{tab:baselines}}
\end{table}

\subsection{Single Shift Synchronization For LRS2-BBC}

In order to create the evaluation set for this experiment, we uniformly shifted the audio of LRS2's test set, by a random shift of up to one second. We compare our method to SyncNet and the modified DTA method (matching is performed in the other direction). Despite some effort, we could not make the original DTA method work well on this dataset. SyncNet is run in the accurate mode, in which it is applied multiple times at different strides followed by a voting procedure. Our method is run only once.


The results are reported in Tab.~\ref{tab:ssp_table}. As  can be seen, our method outperforms the other methods by a significant margin. In 88\% of 1,200 videos, our prediction was exact. Moreover, the error for our method was never more than one frame. The results of our method without the attention mechanism are added to the table in order to study the importance of this module. As can be seen, the performance without the attention mechanism is not competitive.

\begin{table}
\begin{center}
\begin{tabular}{p{2.5cm}p{1cm}p{1cm}p{2.5cm}}  
    \toprule
    Method & Mean shift error (frames) & Max shift error (frames)  & Top-1 accuracy per video \\
    \midrule
    SyncNet & 0.45 & 36 & 60\%  \\
    Modified DTA &  0.25 & 27 & 83\%\\ 
    Ours &  \textbf{0.12} & \textbf{1} & \textbf{88\%}\\
    \midrule
    Ours (no Att.) & 0.61 & 13 & 45\% \\
  \bottomrule
  \end{tabular}
\end{center}
\caption{Single shift performance evaluation for LRS2.  The top-1 accuracy per video indicates the percentage of videos where the global shift was predicted exactly. \label{tab:ssp_table}}
\end{table}


\subsection{Dynamic Per-Frame Shifts For LRS2-BBC}

As detailed in Sec.~\ref{Datasets}, using the same pipeline that was used for creating the training set, an evaluation set for LRS2-BBC was created.

The results for the modified DTA method as well as for our method are reported in Tab.~\ref{tab:spf_lrs}.  The original DTA method, which is the only other local alignment method, cannot be applied here since the video is modified and the audio remains unchanged. As can be seen,  our method greatly outperforms the baseline method in both the average shift error and in the percent of frames that were assigned the correct alignment. The table also contains the results of our method without the attention mechanism. The results for this method are considerably worse than those of our complete method.

\begin{table}
\begin{center}
\begin{tabular}{@{}lc@{~~}c@{}}
    \toprule
    Method & Mean shift error~(frames) & Top-1 accuracy\\
    \midrule
    Modified DTA & 1.94 & 25\% \\
     Ours  & \textbf{0.85} & \textbf{60\%} \\
     \midrule
       Ours (no Att.) & 1.60 & 26\% \\
 
  \bottomrule
  \end{tabular}
\end{center}
\caption{Results on the LRS2-BBC benchmark, where shifts vary locally. We have measured the average shift error in frames and the percentage of frames where the local shift was predicted exactly (Accuracy). \label{tab:spf_lrs}}
\end{table}

\subsection{Shift Per-Frame Synchronization For TCD-TIMIT}

For the TCD-TIMIT benchmark, the videos for sentences sa1 and sa2 of the first 5 speakers were used. All permutations between source and destination speakers, for each of the two sentences, were evaluated, resulting in 40 different experiments.

In each experiment, we have the video of speaker $a$ aligned to the audio of speaker $b$. In order to evaluate performance, we can compare the modified video of speaker $a$ to the video of speaker $b$. We perform this comparison by considering the correlation of facial keypoints in those videos.



The OpenFace~\cite{Baltrusaitis2018OpenFace2F} software is used to extract these keypoints, and we calculate the Pearson correlation in the mouth height (measured as the vertical distance between key points 63 and 67) 
and in the mouth width (horizontal distance between facial key points 61 and 65). 
A high correlation indicates better alignment. 

We compare the correlations along both the $x$ and $y$ axes with no alignment, as well as the modified DTA method (since the video is modified) and our method. 
The results, listed in Tab.~\ref{tab:kp_tab}, indicate that the proposed method outperforms the modified DTA method. 

In addition to measuring the correlation between the aligned video and the original video that is associated with the voice, the quality of the generated video also needs to be evaluated. When a video is generated by duplicating or deleting frames, it is desirable to duplicate and delete as few frames as possible. In addition, we would prefer situations in which consecutive frames in the original video remain consecutive in the modified version of it. Lastly, we would like to cover as many of the original frames, i.e., have as many unique frames as possible present in the modified video. As Tab.~\ref{tab:stats_tab} shows, our method has a sizable advantage in all four criteria (number of duplicates, number of deletions, number of consecutive frames copied sequentially, and number of unique frames in the modified video).

\begin{table}
\begin{center}
\begin{tabular}{lcc}
    \toprule
    Method & corr x & corr y \\
    \midrule
    No alignment & 0.31 & 0.46 \\
    Modified DTA & 0.40 & 0.47 \\
    Ours & \textbf{0.42} & \textbf{0.50} \\
  \bottomrule
  \end{tabular}
\end{center}
\caption{Average Pearson correlation for mouth height (correlation y) and width (correlation x) on the TCD-TIMIT cross speaker benchmark. \label{tab:kp_tab}}
\end{table}

\begin{table}
\begin{center}
\begin{tabular}{lcccc}
    \toprule
    Method & dup$\downarrow$ & del$\downarrow$ & conseq$\uparrow$& unique$\uparrow$ \\
    \midrule
    Modified DTA & 67.10 & 80.25 & 20.48 & 40.85 \\
    Ours & \textbf{38.45} & \textbf{44.85} & \textbf{58.95} & \textbf{76.25} \\
  \bottomrule
  \end{tabular}
\end{center}
\caption{Number of deleted and duplicated frames on the TCD-TIMIT benchmark, as well as the number of frames in which the matching frame is the consecutive frame of the match of the previous frame and the number of unique frames. \label{tab:stats_tab}}
\end{table}




In addition, we have also repeated the experiment such that the audio is aligned to the video, as is done in the unmodified DTA method. For our method, this required modifying it to perform the alignment in the other direction. For this purpose, we replace the roles of audio and video and let each distance vector $\Vec{\rho_k}$ represent the distances of 75 audio frames to a single video frame, out of 25 such frames. We do this without retraining the network, assuming symmetry in the alignment process. Additional performance may be gained by training specifically for this scenario.

For evaluation, we need to compare the warped audio of speaker $b$ to the original audio of speaker $a$. This is done using the Mel Cepstral Distortion (MCD) measure. The closer the sequences are, the lower the MCD score. A second variant of this distortion (MCD-DTW), employs Dynamic Time Warping to minimize the MCD score. In the context of alignment methods, MCD-DTW measures the extent to which content is maintained during the alignment process. In other words, one can obtain a relatively low MCD score, while removing part of the original audio, instead of mapping it. The MCD-DTW score performs its own alignment and is more sensitive to such neglect. 

For a fair comparison, we normalized the audio of the reference and warped inputs for all methods into -30 dBFS. The Mel generalized cepstral analysis is done with a frame length of 64 ms and hop length of 5 ms, in order to extract 60 length features for the warped and reference audio streams. 
The obtained scores are reported in Table \ref{tab:mcd_tabl.}. Our method outperforms DTA with respect to both MCD and MCD-DTW. 

\begin{table}
\begin{center}
\begin{tabular}{ccc}
    \toprule
    Method & MCD & MCD-DTW\\
    \midrule
    DTA & 10.37 $\pm$ 1.84 & 5.64 $\pm$ 1.19 \\
    Ours (audio warp version) & \textbf{9.13 $\pm$ 2.11} & \textbf{3.92 $\pm$ 1.48} \\
  \bottomrule
  \end{tabular}
\end{center}
\caption{MCD and MCD DTW scores for TCD-TIMIT (Mean $\pm$ SD; lower is better) \label{tab:mcd_tabl.}}
\end{table}

\subsection{User Study}

In order to quantitatively evaluate the performance of our method, we conducted a user study. TCD-TIMIT dataset was used in order to synthesize eight different videos, where the audio of a source speaker was synchronized with the video of a different target speaker. All pairs are both either males or females and
the videos consist of seven different speakers, four males and three females, saying sentences sa1 and sa2.

The participants were presented with the videos and were asked to tag each video with a real or a modified label. The participants were told that we may present any combination of real and modified videos, while, in reality, we always presented two real and two modified videos that were randomly selected and ordered. 

The participants were given ample time, were able to watch the videos an unlimited number of times, and were challenged to do their best. The videos were watched on a 27 inch computer monitor, and were displayed at a resolution of $1920 \times 1080$. Ten males and ten females between the ages of 20--50 participated in the study. The results are shown in Tab.~\ref{tab:user_study}. As can be seen, the results clearly demonstrate that most of the time participants are not able to distinguish between real and modified videos, which indicates the artifact-free nature of the modified video and that these videos are extremely well-synchronized.

\begin{table}
\begin{center}
\begin{tabular}{ccc}
  \toprule

Answer  &    Real video & Modified video    \\
    \midrule
``The video is real'' &    $86\%$ & $83\%$\\
``The video is fake'' &    $14\%$ & $17\%$\\
    \bottomrule
  
  \end{tabular}
\end{center}
\caption{User study results. The 20 users were asked to specify for a set of four videos, which ones are real and which are fake. \label{tab:user_study}}
\end{table}

\subsection{Application to TTS Voices}

Our method allows synchronizing the audio and the video streams of two different speakers, where the tempo of the two streams is completely different. One potential usage is for same-language dubbing, which is a standard industry practice known as looping. Same-language dubbing is required for cases where filming special effects shots or in the presence of a background noise. Another common case is with shows aimed at preschoolers where dubbing is done in order to ensure the usage of a "right accent", as they learn to speak their native tongue. Another use case are animation films and musicals, where the actor's singing ability may not be good enough, and may be dubbed by a professional singer.

In order to illustrate the capabilities of our method for same-language dubbing, we use a text to speech module in order to dub variety of speakers. We use free online services as the text to speech system, and a variety of YouTube videos, including hosting shows, Obama's weekly addresses, etc. as the video stream. The results are very convincing and are available as part of the supplementary videos.


\subsection{Runtime}
The proposed method and the modified DTA baseline are both running significantly faster than real-time. Our method is $1.7$ times faster than the modified DTA baseline, once the input tensors have been prepared. However, the run time of both is negligible in comparison to the time it takes to extract the facial keypoints. 

  


\section{Conclusions and Future Work}

We present a novel method for lip syncing. The method obtains a sizable gap in accuracy over the literature on multiple benchmarks. Applying the method, even in challenging scenarios where the audio and the video do not perfectly match, results in an artifact-free, well-aligned video. As we demonstrate, users find it hard to distinguish, based on either quality or the accuracy of the audio-video alignment, between the modified video and the original unmodified video. While we have demonstrated compelling results, further improvements may be considered.
In order to eliminate the post-processing step, 
a monotonic attention mechanism \cite{2017arXiv170400784R}, \cite{2017arXiv171205382C}, \cite{2019arXiv190912406M},
may be utilized, in order to force the decoder to produce a monotonically increasing alignment.
In addition, monotonic constraints could be added to the loss function, as well as using an Euclidean or an ordinal loss function, in order to weight the magnitude of the prediction error. 
Our code and scripts for training and running the method are attached as supplementary and would be released as open source.

\section{Acknowledgements}

This project has received funding from the European Research Council (ERC) under the European
Union’s Horizon 2020 research and innovation programme (grant ERC CoG 725974).

{\small
\bibliographystyle{ieee}
\bibliography{main}

\begin{thebibliography}{1}\itemsep=-1pt

\bibitem{Alpher02}
A.~Alpher.
\newblock Frobnication.
\newblock {\em Journal of Foo}, 12(1):234--778, 2002.

\bibitem{Alpher03}
A.~Alpher and J.~P.~N. Fotheringham-Smythe.
\newblock Frobnication revisited.
\newblock {\em Journal of Foo}, 13(1):234--778, 2003.

\bibitem{Alpher04}
A.~Alpher, J.~P.~N. Fotheringham-Smythe, and G.~Gamow.
\newblock Can a machine frobnicate?
\newblock {\em Journal of Foo}, 14(1):234--778, 2004.

\bibitem{Authors06b}
Authors.
\newblock Frobnication tutorial, 2006.
\newblock Supplied as additional material {\tt tr.pdf}.

\bibitem{Authors06}
Authors.
\newblock The frobnicatable foo filter, 2011.
\newblock Face and Gesture submission ID 324. Supplied as additional material
  {\tt fg324.pdf}.

\end{thebibliography}


\begin{thebibliography}{10}\itemsep=-1pt

\bibitem{MULER}
{\em Dynamic Time Warping}, pages 69--84.
\newblock Springer Berlin Heidelberg, Berlin, Heidelberg, 2007.

\bibitem{2018arXiv180902108A}
T.~{Afouras}, J.~{Son Chung}, A.~{Senior}, O.~{Vinyals}, and A.~{Zisserman}.
\newblock {Deep Audio-Visual Speech Recognition}.
\newblock {\em arXiv e-prints}, page arXiv:1809.02108, Sep 2018.

\bibitem{anguera2014audio}
X.~Anguera, J.~Luque, and C.~Gracia.
\newblock Audio-to-text alignment for speech recognition with very limited
  resources.
\newblock In {\em Fifteenth Annual Conference of the International Speech
  Communication Association}, 2014.

\bibitem{Argones}
E.~Argones~Rúa, H.~Bredin, C.~García-Mateo, G.~Chollet, and D.~Jiménez.
\newblock Audio-visual speech asynchrony detection using co-inertia analysis
  and coupled hidden markov models.
\newblock {\em Pattern Anal. Appl.}, 12:271--284, 09 2009.

\bibitem{2014arXiv1409.0473B}
D.~Bahdanau, K.~Cho, and Y.~Bengio.
\newblock Neural machine translation by jointly learning to align and
  translate.
\newblock {\em arXiv preprint arXiv:1409.0473}, 2014.

\bibitem{Baltrusaitis2018OpenFace2F}
T.~Baltrusaitis, A.~B. Zadeh, Y.~C. Lim, and L.-P. Morency.
\newblock Openface 2.0: Facial behavior analysis toolkit.
\newblock {\em 2018 13th IEEE International Conference on Automatic Face \&
  Gesture Recognition (FG 2018)}, pages 59--66, 2018.

\bibitem{2017arXiv171205382C}
C.-C. {Chiu} and C.~{Raffel}.
\newblock {Monotonic Chunkwise Attention}.
\newblock {\em arXiv e-prints}, page arXiv:1712.05382, Dec 2017.

\bibitem{Chung2016LipRI}
J.~S. Chung and A.~Zisserman.
\newblock Lip reading in the wild.
\newblock In {\em ACCV}, 2016.

\bibitem{Chung2016OutOT}
J.~S. Chung and A.~Zisserman.
\newblock Out of time: Automated lip sync in the wild.
\newblock In {\em ACCV Workshops}, 2016.

\bibitem{2018arXiv180908001C}
S.-W. {Chung}, J.~{Son Chung}, and H.-G. {Kang}.
\newblock {Perfect match: Improved cross-modal embeddings for audio-visual
  synchronisation}.
\newblock {\em arXiv e-prints}, page arXiv:1809.08001, Sep 2018.

\bibitem{pmlr-v9-glorot10a}
X.~Glorot and Y.~Bengio.
\newblock Understanding the difficulty of training deep feedforward neural
  networks.
\newblock In Y.~W. Teh and M.~Titterington, editors, {\em Proceedings of the
  Thirteenth International Conference on Artificial Intelligence and
  Statistics}, volume~9 of {\em Proceedings of Machine Learning Research},
  pages 249--256, Chia Laguna Resort, Sardinia, Italy, 13--15 May 2010. PMLR.

\bibitem{2018arXiv180806250H}
T.~{Halperin}, A.~{Ephrat}, and S.~{Peleg}.
\newblock {Dynamic Temporal Alignment of Speech to Lips}.
\newblock {\em arXiv e-prints}, page arXiv:1808.06250, Aug 2018.

\bibitem{7050271}
N.~{Harte} and E.~{Gillen}.
\newblock Tcd-timit: An audio-visual corpus of continuous speech.
\newblock {\em IEEE Transactions on Multimedia}, 17(5):603--615, May 2015.

\bibitem{haubold2007alignment}
A.~Haubold and J.~R. Kender.
\newblock Alignment of speech to highly imperfect text transcriptions.
\newblock In {\em 2007 IEEE International Conference on Multimedia and Expo},
  pages 224--227. IEEE, 2007.

\bibitem{NVIDIA}
T.~Karras, T.~Aila, S.~Laine, A.~Herva, and J.~Lehtinen.
\newblock Audio-driven facial animation by joint end-to-end learning of pose
  and emotion.
\newblock {\em ACM Transactions on Graphics}, 36:1--12, 07 2017.

\bibitem{2014arXiv1412.6980K}
D.~P. {Kingma} and J.~{Ba}.
\newblock {Adam: A Method for Stochastic Optimization}.
\newblock {\em arXiv e-prints}, page arXiv:1412.6980, Dec 2014.

\bibitem{10.2307/2030214}
J.~B. Kruskal.
\newblock An overview of sequence comparison: Time warps, string edits, and
  macromolecules.
\newblock {\em SIAM Review}, 25(2):201--237, 1983.

\bibitem{2018arXiv180101442K}
R.~{Kumar}, J.~{Sotelo}, K.~{Kumar}, A.~{de Brebisson}, and Y.~{Bengio}.
\newblock {ObamaNet: Photo-realistic lip-sync from text}.
\newblock {\em arXiv e-prints}, page arXiv:1801.01442, Dec 2017.

\bibitem{doi:10.1002/vis.4340020404}
J.~Lewis.
\newblock Automated lip-sync: Background and techniques.
\newblock {\em The Journal of Visualization and Computer Animation},
  2(4):118--122, 1991.

\bibitem{2019arXiv190912406M}
X.~{Ma}, J.~{Pino}, J.~{Cross}, L.~{Puzon}, and J.~{Gu}.
\newblock {Monotonic Multihead Attention}.
\newblock {\em arXiv e-prints}, page arXiv:1909.12406, Sep 2019.

\bibitem{Marcheret2015DetectingAS}
E.~Marcheret, G.~Potamianos, J.~Vopicka, and V.~Goel.
\newblock Detecting audio-visual synchrony using deep neural networks.
\newblock In {\em INTERSPEECH}, 2015.

\bibitem{5745053}
S.~{Morishima}, S.~{Ogata}, K.~{Murai}, and S.~{Nakamura}.
\newblock Audio-visual speech translation with automatic lip syncqronization
  and face tracking based on 3-d head model.
\newblock In {\em 2002 IEEE International Conference on Acoustics, Speech, and
  Signal Processing}, volume~2, pages II--2117--II--2120, May 2002.

\bibitem{2017arXiv170400784R}
C.~{Raffel}, M.-T. {Luong}, P.~J. {Liu}, R.~J. {Weiss}, and D.~{Eck}.
\newblock {Online and Linear-Time Attention by Enforcing Monotonic Alignments}.
\newblock {\em arXiv e-prints}, page arXiv:1704.00784, Apr 2017.

\bibitem{4351913}
M.~E. {Sargin}, Y.~{Yemez}, E.~{Erzin}, and A.~M. {Tekalp}.
\newblock Audiovisual synchronization and fusion using canonical correlation
  analysis.
\newblock {\em IEEE Transactions on Multimedia}, 9(7):1396--1403, Nov 2007.

\bibitem{2013arXiv1312.6120S}
A.~M. {Saxe}, J.~L. {McClelland}, and S.~{Ganguli}.
\newblock {Exact solutions to the nonlinear dynamics of learning in deep linear
  neural networks}.
\newblock {\em arXiv e-prints}, page arXiv:1312.6120, Dec 2013.

\bibitem{6126545}
S.~{Shariat} and V.~{Pavlovic}.
\newblock Isotonic cca for sequence alignment and activity recognition.
\newblock In {\em 2011 International Conference on Computer Vision}, pages
  2572--2578, Nov 2011.

\bibitem{2017arXiv170502966S}
J.~{Son Chung}, A.~{Jamaludin}, and A.~{Zisserman}.
\newblock {You said that?}
\newblock {\em arXiv e-prints}, page arXiv:1705.02966, May 2017.

\bibitem{Suwajanakorn:2017:SOL:3072959.3073640}
S.~Suwajanakorn, S.~M. Seitz, and I.~Kemelmacher-Shlizerman.
\newblock Synthesizing obama: Learning lip sync from audio.
\newblock {\em ACM Trans. Graph.}, 36(4):95:1--95:13, July 2017.

\bibitem{trigeorgis2016deep}
G.~Trigeorgis, M.~A. Nicolaou, S.~Zafeiriou, and B.~W. Schuller.
\newblock Deep canonical time warping.
\newblock In {\em Proceedings of the IEEE Conference on Computer Vision and
  Pattern Recognition}, pages 5110--5118, 2016.

\bibitem{2014arXiv1411.5726V}
R.~{Vedantam}, C.~L. {Zitnick}, and D.~{Parikh}.
\newblock {CIDEr: Consensus-based Image Description Evaluation}.
\newblock {\em arXiv e-prints}, page arXiv:1411.5726, Nov 2014.

\bibitem{2016arXiv161108657Z}
A.~{Zadeh}, T.~{Baltru{\v{s}}aitis}, and L.-P. {Morency}.
\newblock {Convolutional Experts Constrained Local Model for Facial Landmark
  Detection}.
\newblock {\em arXiv e-prints}, page arXiv:1611.08657, Nov 2016.

\bibitem{zhou2012generalized}
F.~Zhou and F.~De~la Torre.
\newblock Generalized time warping for multi-modal alignment of human motion.
\newblock In {\em 2012 IEEE Conference on Computer Vision and Pattern
  Recognition}, pages 1282--1289. IEEE, 2012.

\bibitem{zhou2009canonical}
F.~Zhou and F.~Torre.
\newblock Canonical time warping for alignment of human behavior.
\newblock In {\em Advances in neural information processing systems}, pages
  2286--2294, 2009.

\bibitem{zhu2015aligning}
Y.~Zhu, R.~Kiros, R.~Zemel, R.~Salakhutdinov, R.~Urtasun, A.~Torralba, and
  S.~Fidler.
\newblock Aligning books and movies: Towards story-like visual explanations by
  watching movies and reading books.
\newblock In {\em Proceedings of the IEEE international conference on computer
  vision}, pages 19--27, 2015.

\bibitem{Zoric}
G.~Zoric and S.~Igor.
\newblock A real-time lip sync system using a genetic algorithm for automatic
  neural network configuration.
\newblock pages 1366--1369, 01 2005.

\end{thebibliography}
}

\end{document}